\documentclass{article} 
\usepackage{nips15submit_e,times}
\usepackage{url}
\usepackage{graphicx}
\usepackage{subfig}
\usepackage{import}
\usepackage{algorithm}
\usepackage{algorithmic}
\usepackage{amsfonts}
\usepackage{mathtools}
\usepackage{array}
\usepackage{longtable}
\usepackage{sidecap}

\usepackage[numbers]{natbib}

\title{Memory-based control with recurrent neural networks}

\author{
Nicolas Heess* \hspace{3mm}
Jonathan J Hunt* \hspace{3mm}
Timothy P Lillicrap \hspace{3mm}
David Silver
\\
Google Deepmind \\
* These authors contributed equally.\\
\texttt{\textit{heess, jjhunt, countzero, davidsilver} @ google.com}
}

\nipsfinalcopy 

\DeclareMathOperator{\E}{\mathbb{E}} 

\definecolor{gray}{rgb}{0.7,0.7,0.7}

\newcommand{\partialAt}[2]{\frac{\partial #1}{\partial #2}}
\newcommand{\expectationE}[2]{ \mathbb{E}_{#2}  \left[ #1 \right] }

\begin{document}

\maketitle

\begin{abstract}
Partially observed control problems are a challenging aspect of
reinforcement learning. We extend two related, model-free algorithms for
continuous control -- deterministic policy gradient and stochastic value gradient
-- to solve partially observed domains using recurrent neural networks trained with
backpropagation through time.
We demonstrate that this approach, coupled with long-short term memory is able
to solve a variety of physical control problems exhibiting an
assortment of memory requirements. These include the short-term
integration of information from noisy sensors and the identification of system
parameters, as well as long-term memory problems that require preserving
information over many time steps. We also demonstrate success on  a
combined exploration and memory problem in the form of a simplified version of
the well-known Morris water maze task.  Finally, we show that our approach can
deal with high-dimensional observations by learning directly from
pixels.
We find that recurrent deterministic and stochastic policies are able to
learn similarly good solutions to these tasks, including the water maze where
the agent must learn effective search strategies.
\end{abstract}

\section{Introduction}

The use of neural networks for solving continuous control problems has a long
tradition. Several recent papers successfully apply model-free,
direct policy search methods to the problem of learning neural network control
policies for challenging continuous domains with many degrees of freedoms
\cite{balduzzi2015compatible, heess2015svg,
      lillicrap2015continuous,schulman2015trust,schulman2015advantage,levine2015end}.
However, all of this work assumes fully observed state.

Many real world control problems are
partially observed. Partial observability can arise from different sources
including the need to remember information that is only temporarily available
such as a way sign in a navigation task, sensor limitations or noise, unobserved
variations of the plant under control (system identification), or state-aliasing
due to function approximation. Partial observability also arises naturally in
many tasks that involve control from vision: a static image of a dynamic scene
provides no information about velocities, occlusions occur as a consequence of
the three-dimensional nature of the world, and most vision sensors are
bandwidth-limited and only have a restricted field-of-view.

Resolution of partial observability is non-trivial. Existing methods
can roughly be divided into two broad classes:

On the one hand there are approaches that
explicitly maintain a belief state that corresponds to the
distribution over the world state given the observations so far. This approach
has two major disadvantages: The first is the need for a model, and the second
is the computational cost that is typically associated with the update of the
belief state \cite{kaelbling1998planning, shani2013survey}.

On the other hand there are model free approaches that learn to form
memories based on interactions with the world. This is challenging
since it is a priori unknown which features of the observations will be
relevant later, and associations may have to be formed over many steps.
For this reason, most model free approaches tend to assume the fully-observed
case.
In practice, partial observability is often solved by hand-crafting
a solution such as providing multiple-frames at each timestep
to allow  velocity estimation
\cite{mnih2015human, lillicrap2015continuous}.

In this work we investigate a natural extension of
two recent, closely related policy gradient algorithms
for learning continuous-action policies to handle partially observed problems.
We primarily consider the Deterministic Policy Gradient algorithm (DPG) \cite{silver2014deterministic}, which
is an off-policy policy gradient algorithm that has recently produced promising
results on a broad range of difficult, high-dimensional continuous control
problems, including direct control from pixels \cite{lillicrap2015continuous}.
DPG is an actor-critic algorithm that uses a learned approximation of the action-value (Q) function to obtain
approximate action-value gradients. These are then used
to update a deterministic policy via the chain-rule. We also consider DPG's
stochastic counterpart, SVG(0) (\cite{heess2015svg}; SVG stands for ``Stochastic
Value Gradients'') which similarly updates the policy via backpropagation of
action-value gradients from an action-value critic but learns a
stochastic policy.

We modify both algorithms to use
recurrent networks trained with backpropagation through time.
We demonstrate that the resulting algorithms, Recurrent
DPG (RDPG) and Recurrent SVG(0) (RSVG(0)), can be applied to a number of
partially observed physical control problems with diverse memory
requirements. These problems include: short-term integration of sensor
information to estimate the system state (pendulum and cartpole
swing-up tasks without velocity
information); system identification (cart pole swing-up with
variable and unknown pole-length); long-term memory (a robot arm
that needs to reach out and grab a payload to move it to the position the arm
started from); as well as a simplified version of the water maze task which
requires the agent to learn an exploration strategy to find a hidden platform
and then remember the platform's position in order to return to it subsequently.
We also demonstrate successful control directly from pixels.

Our results suggest that actor-critic algorithms that rely on bootstrapping for
estimating the value function can be a viable option for learning control
policies in partially observed domains. We further find that, at least in the
setup considered here, there is little performance difference between stochastic
and deterministic policies, despite the former being typically presumed to be
preferable in partially observed domains.


\section{Background}

We model our environment as discrete-time,
partially-observed Markov Decision process (POMDP). A POMDP is described
a set of environment states $\mathcal{S}$ and a set of actions $\mathcal{A}$,
an initial state distribution $p_0(s_0)$,
a transition function $p(s_{t + 1} | s_{t}, a_{t})$ and reward function
$r(s_t, a_t)$. This underlying MDP is partially observed when the agent
is unable to observe the state $s_t$ directly and instead receives observations
from the set $\mathcal{O}$ which are conditioned on the underlying state
$p(o_t | s_{t} )$.

The agent only indirectly observes the underlying state of the MDP
through the observations. An optimal agent may, in principle, require
access to the entire history
$h_t = (o_1, a_1, o_2, a_2, ... a_{t-1}, o_t)$.

%
The goal of the agent is thus to learn a policy $\pi(h_t)$ which maps from the
history to a distribution over actions $P(\mathcal{A})$ which
maximizes the expected discounted reward (below we consider both stochastic and
deterministic policies).
%
%
%
%
For stochastic policies we want to maximise
\begin{equation}
  J = \E_{\tau}\left[ \sum_{t=1}^\infty \gamma^{t - 1} r(s_t, a_t)  \right],
\end{equation}
where the trajectories $\tau = (s_1, o_1, a_1, s_2, \dots )$ are drawn from the trajectory distribution induced by the policy $\pi$: $p(s_1)p(o_1 | s_1) \pi(a_1|h_1) p(s_2 | s_1,a_1) p(o_2 | s_2 )\pi(a_2 | h_2) \dots$ and where $h_t$ is defined as above.
For deterministic policies we replace $\pi$ with a deterministic function
$\mu$ which maps directly from states $\mathcal{S}$ to actions $\mathcal{A}$ and
 we replace $a_t\sim \pi(\cdot | h_t) $
with $a_t = \mu(h_t)$.

In the algorithms below we make use of the action-value function $Q^\pi$.
For a fully observed MDP, when we have access to $s$, the action-value function is defined as the expected future discounted
reward when in state $s_t$ the agent takes action $a_t$ and thereafter follows  policy $\pi$.
Since we are interested in the partially observed case where the agent does not
have access to $s$ we instead define $Q^\pi$ in terms of $h$:
\begin{equation}
Q^\pi(h_t, a_t) = \E_{s_t | h_t} \left [ r_t(s_t,a_t) \right ]+ \E_{ \tau_{>t} |
h_t, a_t } \left [ \sum_{i=1}^\infty \gamma^{i} r(s_{t + i}, a_{t +i} ) \right]
\end{equation}
where $\tau_{> t} = (s_{t+1}, o_{t+1}, a_{t+1} \dots)$ is the
future trajectory and the two expectations are taken with respect to the
conditionals $p(s_t | h_t)$ and $p(\tau_{> t} | h_t, a_t)$ of the trajectory
distribution associated with $\pi$. Note that this equivalent to defining $Q^\pi$ in terms of the
belief state since $h$ is a sufficient statistic.

Obviously, for most POMDPs of interest, it is not tractable to condition on the
entire sequence of observations. A central challenge is to learn how to summarize
the past in a scalable way.

%


\section{Algorithms}

\subsection{Recurrent DPG} We extend the  Deterministic Policy Gradient (DPG)
algorithm for MDPs introduced in \cite{silver2014deterministic} to deal with
partially observed domains and pixels.
The core idea of the DPG algorithm for the \emph{fully observed} case is that for a deterministic policy
$\mu^\theta$ with parameters $\theta$, and given access to the true action-value
function associated with the current policy $Q^\mu$, the policy can be updated
by backpropagation:
\begin{equation}
\partialAt{J(\theta)}{\theta} =
\expectationE{ 	  \left. \partialAt{ Q^\mu(s,a)}{a} \right \rvert_{a = \mu^\theta(s)} \partialAt{\mu^\theta(s)}{\theta}  } 
{s \sim \rho^\mu},
\label{eq:DPGpolicyUpdate}
\end{equation}
where the expectation is taken with respect to the (discounted) state visitation distribution $\rho^\mu$ induced by the
current policy $\mu^\theta$ \cite{silver2014deterministic}. Similar ideas had previously been exploited in
NFQCA \cite{hafner2011reinforcement} and in the ADP \cite{lewis2009reinforcement}
community.
In practice the exact action-value function $Q^\mu$  is replaced by an
approximate (critic) $Q^\omega$ with parameters $\omega$ that is
differentiable in $a$ and which can be learned e.g.\ with Q-learning.

In order to ensure the applicability of our approach to
large observation spaces (e.g.\ from pixels), we use
neural networks for all function approximators. These networks, with
convolutional layers have proven effective at many sensory processing tasks
\cite{krizhevsky2012imagenet, razavian2014cnn},
and been demonstrated to be effective for scaling reinforcement learning to
large state spaces \cite{lillicrap2015continuous,mnih2015human}.
\cite{lillicrap2015continuous} proposed modifications to DPG necessary in order
to learn effectively with deep neural networks which we make use of here (cf.\
sections \ref{sec:Algorithms:ER}, \ref{sec:Algorithms:Target}).

Under partial observability the optimal policy and the associated action-value
function are both functions of the entire preceding observation-action history
$h_t$. The primary change we introduce is
the use of recurrent neural networks, rather than feedforward networks,
in order to allow the network to learn to preserve (limited) information about
the past which is needed in order to solve the POMDP. Thus, writing $\mu(h)$ and
$Q(h,a)$ rather than $\mu(s)$ and $Q(s,a)$ we obtain the following policy
update:

\begin{equation}
\partialAt{J(\theta)}{\theta} =
\expectationE{
	\sum_t \gamma^{t-1} \left. \partialAt{ Q^\mu(h_t,a)}{a} \right \rvert_{a = \mu^\theta(h_t)} \partialAt{\mu^\theta(h_t)}{\theta}
}
{\tau},
\label{eq:RDPGpolicyUpdate}
\end{equation}
where we have written the expectation now explicitly over entire trajectories $\tau = (s_1, o_1, a_1,
s_2, o_2, a_2, \dots )$ which are drawn from the trajectory distribution induced by the current policy
and $h_t = (o_1, a_1, \dots, o_{t-1}, a_{t-1}, o_t) $ is the observation-action trajectory prefix at time step $t$, both as introduced above\footnote{ A
discount factor $\gamma^t$ appears implicitly in the update which is absorbed in
the discounted state-visitation distribution in eq.\ \ref{eq:DPGpolicyUpdate}.
In practice we ignore this term as is often done in policy gradient
implementations in practice (e.g.\ \cite{thomas2014bias}). }. In
practice, as in the fully observed case, we replace $Q^\mu$ by learned
approximation $Q^\omega$ (which is also a recurrent
network with parameters $\omega$).
Thus, rather than directly conditioning on the entire observation history,
we effectively train recurrent neural networks to summarize this history in
their recurrent state using backpropagation through time (BPTT).
For long episodes or continuing tasks it is possible to use
truncated BPTT, although we do not use this here.

%
%
%
The full algorithm is given below (Algorithm \ref{algo:rdpg}).

RDPG is an algorithm for learning deterministic policies. As discussed in the
literature \cite{singh94learningwithout,sallans2002reinforcement}
it is possible to construct examples where deterministic
policies perform poorly under partial observability. In RDPG the policy is
conditioned on the entire history but since we are using function approximation
state aliasing may still occur, especially early in
learning. We therefore also investigate a recurrent version of the stochastic
counterpart to DPG: SVG(0) \cite{heess2015svg} (DPG can be seen as the deterministic limit of
SVG(0)). In addition to learning stochastic policies SVG(0) also admits on-policy
learning whereas DPG is inherently off policy (see below).

Similar to
DPG, SVG(0) updates the policy by backpropagation $\partial Q / \partial a$
from the action-value function, but does so for stochastic policies. This is
enabled through a ``re-parameterization'' (e.g.\ \cite{kingma2013autoencoding,rezende2014stochastic})
of the stochastic policy: The
stochastic policy is represented in terms of a fixed, independent noise source
and a parameterized deterministic function that transforms a draw from that
noise source, i.e., in our case, $a = \pi^\theta(h, \nu)$ with $\nu \sim \beta(\cdot)$ where
$\beta$ is some fixed distribution.
For instance, a Gaussian policy $\pi^\theta(a | h) = N(a | \mu^\theta(h), \sigma^2)$ can be re-parameterized as follows:
$a = \pi^\theta(h, \nu) = \mu^\theta(h) + \sigma \nu$ where $\nu \sim N(\cdot | 0,1)$. See \cite{heess2015svg} for more details.

The stochastic policy is updated as follows:
\begin{equation}
\partialAt{J(\theta)}{\theta} =
\expectationE{ 	\sum_t \gamma^{t-1} \left. \partialAt{ Q^{\pi^\theta}(h_t,a)}{a} \right \rvert_{a = \pi^\theta(h_t, \nu_t)} \partialAt{\pi^\theta(h_t,\nu_t)}{\theta}  }{\tau, \nu},
\label{eq:SVGpolicyUpdate}
\end{equation}
with $\tau$ drawn from the trajectory distribution which is conditioned on IID draws of $\nu_t$ from $\beta$ at each time step.
The full algorithm is provided in the supplementary (Algorithm \ref{algo:rsvg0}).

\subsubsection{Off-policy learning and experience replay}
\label{sec:Algorithms:ER}

DPG is typically used in an off-policy setting
due to the fact that the policy is deterministic but exploration is
needed in order to learn the gradient of $Q$ with respect to the actions.
Furthermore, in practice, data efficiency and stability can also be greatly
improved by using experience replay (e.g.\
\cite{hafner2011reinforcement,hausknecht2015deep,lillicrap2015continuous,mnih2015human,heess2015svg})
and we use the same approach here (see Algorithms \ref{algo:rdpg},
\ref{algo:rsvg0}). Thus, during learning we store experienced trajectories in a
database and then replace the expectation in eq.\ (\ref{eq:RDPGpolicyUpdate})
with trajectories sampled from the database.

One consequence of this is a bias in the state distribution in eqs.
(\ref{eq:DPGpolicyUpdate}, \ref{eq:SVGpolicyUpdate}) which no longer corresponds
to the state distribution induced by the current policy . With function
approximation this can lead to a bias in the learned policy, although
this typically ignored in practice.
RDPG and RSVG(0) may similarly be affected;
in fact since policies (and Q) are not just a function of the state but of an entire
action-observation history (eq.\ \ref{eq:RDPGpolicyUpdate}) the bias might be
more severe.

One potential advantage of (R)SVG(0) in this context is that it allows
on-policy learning although we do not explore this possibility
here. We found that off-policy
learning with experience replay remained effective in the partially observed
case.


\subsubsection{Target networks}
\label{sec:Algorithms:Target}

A second algorithmic feature that has been found to greatly improve the
stability of neural-network based reinforcement learning algorithms that rely on
bootstrapping for learning value functions is the use of \emph{target networks}
\cite{hafner2011reinforcement,lillicrap2015continuous,mnih2015human,heess2015svg}:
The algorithm maintains two copies of the value function $Q$ and of the policy
$\pi$ each, with parameters $\theta$ and $\theta'$, and $\omega$ and $\omega'$
respectively. $\theta$ and $\omega$ are the parameters that are being updated by
the algorithm; $\theta'$ and $\omega'$ track them with some delay and are used
to compute the ``targets values'' for the $Q$ function update.
Different authors have explored different approaches to updating $\theta'$ and
$\omega'$. In this work we use ``soft updates'' as in
\cite{lillicrap2015continuous} (see Algorithms \ref{algo:rdpg} and
\ref{algo:rsvg0} below).

\begin{algorithm}[h]
  \caption{RDPG algorithm \label{algo:rdpg}}
  \begin{algorithmic}
    \STATE Initialize critic network $Q^\omega(a_t, h_t )$ and actor $\mu^\theta(h_t )$ with parameters $\omega$ and $\theta$.
    \STATE Initialize target networks $Q^{\omega'}$ and $\mu^{\theta'}$ with weights $\omega' \leftarrow \omega$, $\theta' \leftarrow \theta$.
    \STATE Initialize replay buffer $R$.
    \FOR{episodes = 1, M}
    	\STATE initialize empty history $h_0$
    	\FOR{t = 1, T}
		\STATE receive observation $o_t$
		\STATE $h_t \leftarrow h_{t-1}, a_{t-1}, o_t$ (append observation and previous action to history)
		\STATE select action $a_t = \mu^\theta(h_t) + \epsilon$ (with $\epsilon$: exploration noise)
	\ENDFOR
      	\STATE Store the sequence $(o_1, a_1, r_1 ... o_T, a_T, r_T)$ in $R$
      	\STATE Sample a minibatch of $N$ episodes $(o_1^i, a_1^i, r_1^i, ... o_T^i, a_T^i, r_T^i)_{i=1, \dots ,N}$ from $R$
	\STATE Construct histories $h_t^i = (o_1^i, a_1^i, \dots a_{t-1}^i, o_t^i)$
        \STATE Compute target values for each sample episode
       	$(y_1^i, ... y_T^i)$ using the recurrent target networks
          \begin{equation*}
            y^i_t = r^i_t + \gamma Q^{\omega'}( h^i_{t+1},  \mu^{\theta'}(h^i_{t+1}) )
          \end{equation*}
      \STATE Compute critic update (using BPTT)
      \begin{equation*}
        \Delta \omega = \frac{1}{NT} \sum_i \sum_t \left (y^i_t - Q^\omega(h_t^i, a^i_t ) \right ) \partialAt{Q^\omega(h_t^i, a^i_t )}{\omega}
      \end{equation*}

      \STATE Compute actor update (using BPTT)
      \begin{equation*}
        \Delta \theta = \frac{1}{NT} \sum_i \sum_t \partialAt{Q^\omega(h_t^i, \mu^\theta(h_t^i) )}{a} \partialAt{\mu^\theta(h_t^i)}{\theta}
      \end{equation*}
      \STATE Update actor and critic using Adam \cite{kingma2014adam}
      \STATE Update the target networks
          \begin{align*}
            \omega' & \leftarrow \tau \omega + (1 - \tau) \omega' \\
            \theta'  & \leftarrow \tau \theta + (1 - \tau) \theta'
          \end{align*}
    \ENDFOR
  \end{algorithmic}
\end{algorithm}

\section{Results}

We tested our algorithms on a variety of partial-observed environments,
covering different types of memory problems. Videos of the learned policies for all
the domains are included in our supplementary
videos\footnote{Video of all the learned policies is available at \url{https://youtu.be/V4_vb1D5NNQ}}, we encourage viewing them as
these may provide a better intuition for the environments.
All physical control problems except the simulated water maze
(section \ref{sec:Results:Watermaze}) were simulated in MuJoCo
\cite{todorov2012mujoco}. We tested both standard recurrent networks as well as
LSTM networks.


\subsection{Sensor integration and system identification}

Physical control problems with noisy sensors are one of the
paradigm examples of partially-observed environments.
A large amount of research has focused on how to efficiently integrate
noisy sensory information over multiple timesteps in order to derive accurate
estimates of the system state, or to estimate derivatives of important properties
of the system \cite{thrun2005probabilistic}.

\begin{SCfigure}
  \subfloat[][]{\includegraphics[width=3.5cm]{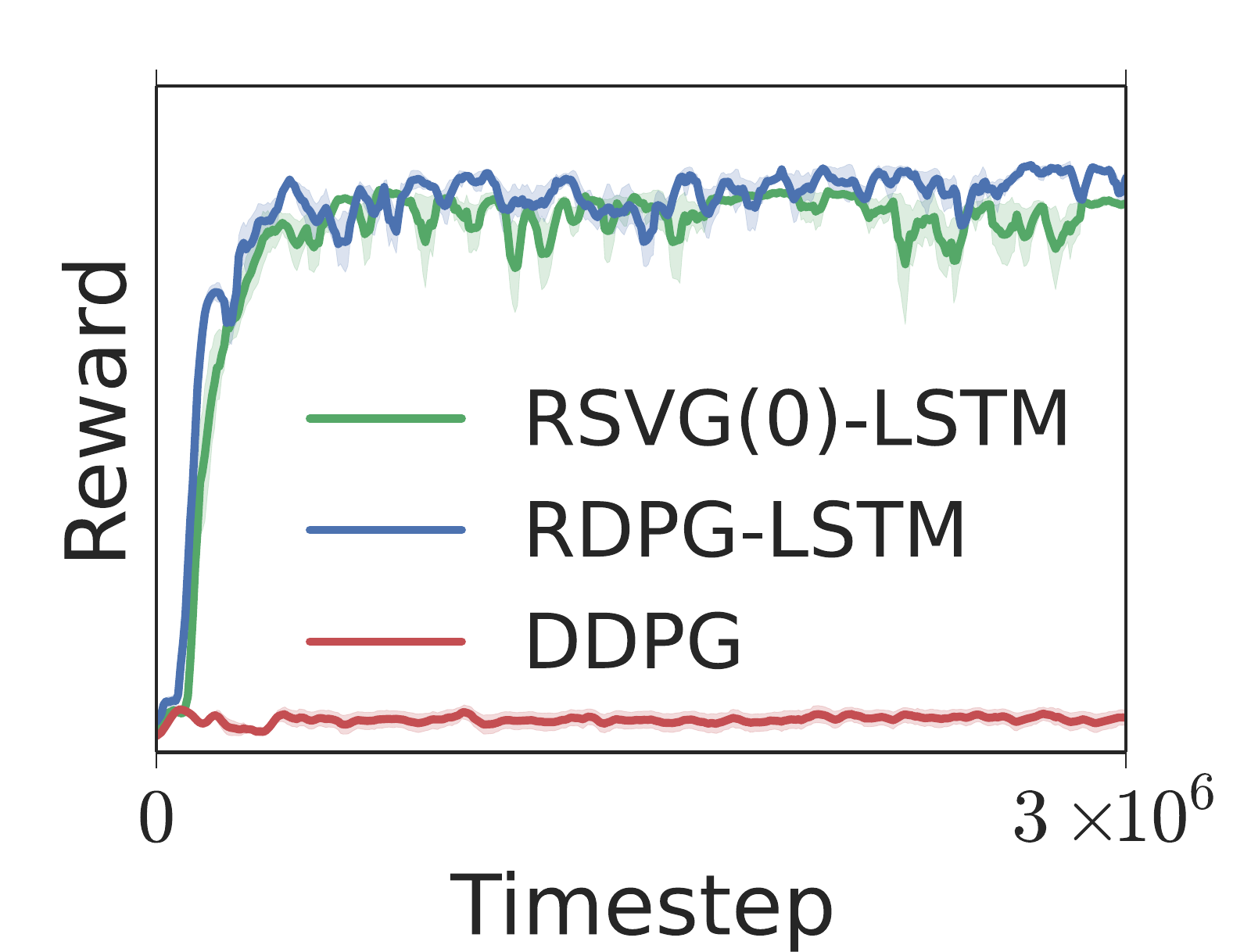} \label{sb:pendulum}}
  \subfloat[][]{\includegraphics[width=3.5cm]{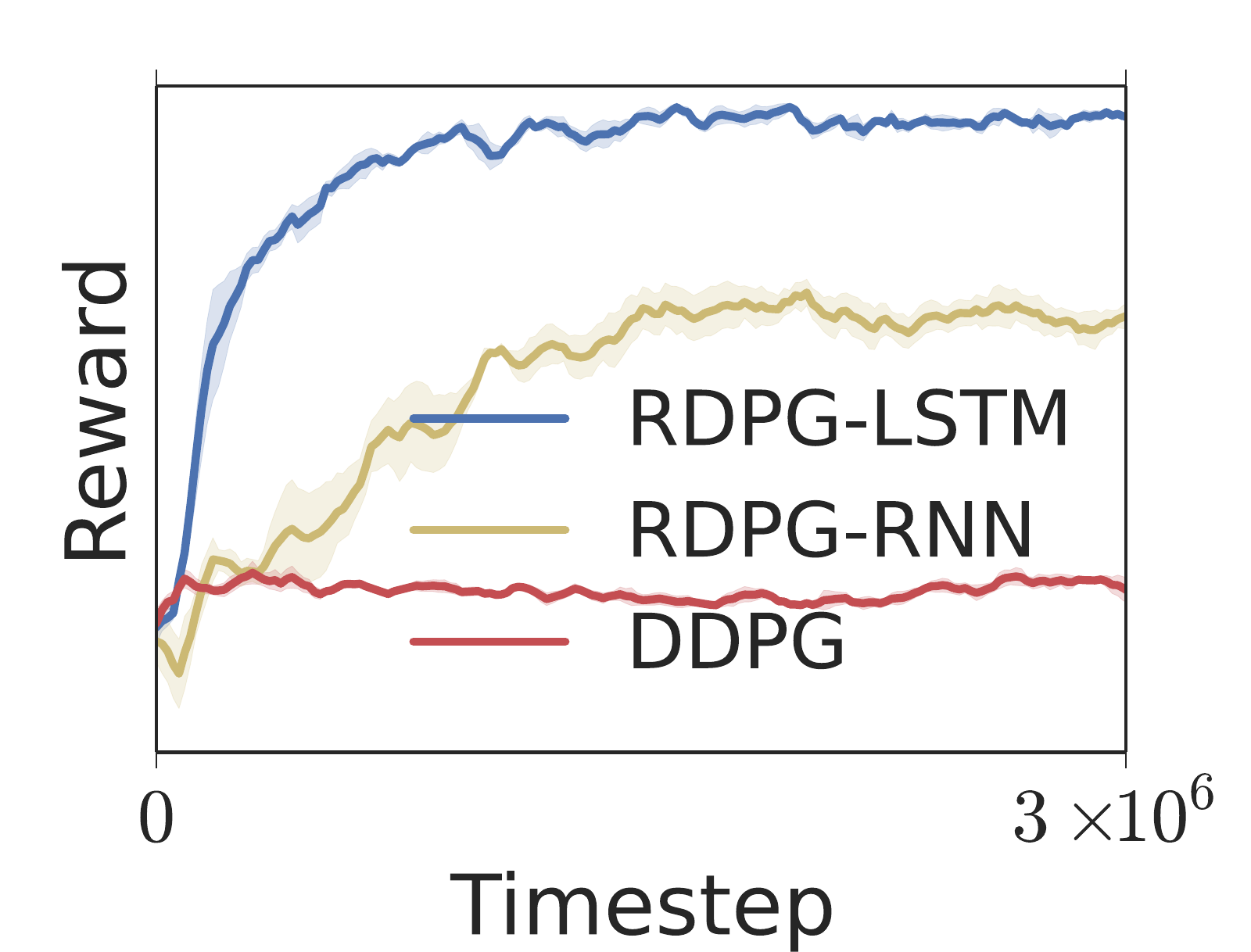} \label{sb:cartpole}}
  \caption{%
  \protect\subref{sb:pendulum} The reward curve for the partially-observed pendulum
  task. Both RDPG and RSVG(0) are able to learn
  policies which bring the pendulum to an upright position.
  \protect\subref{sb:cartpole} The reward curve for the cartpole with no velocity and
  varying cartpole lengths. RDPG with LSTM, is able to reliably learn a good solution for
  this task; a purely feedforward agent (DDPG),
  which will not be able to estimate velocities nor to infer the pole length, is not able
  to solve the problem.
  }%
  \label{fig:pendulum}
\end{SCfigure}

Here, we consider two simple, standard
 control problems often used in reinforcement learning, the under-actuated
 pendulum and cartpole swing up.
We modify these standard benchmarks tasks such that in both cases the agent
receives no direct information of the velocity of any of the components, i.e.\
for the pendulum swing-up task the observation comprises only the angle of the
pendulum, and for cartpole swing-up it is limited to the angle of the pole and
the position of the cart. Velocity is crucial for solving the task
and thus it must be estimated from the history of the system.  Figure
\ref{fig:pendulum}a shows the learning curves for pendulum swing-up. Both RDPG
and RSVG0 were tested on the pendulum task, and are able to learn good solutions
which bring the pole to upright.

For the cartpole swing-up task, in addition to not providing the agent with
velocity information, we also
varied the length of the pole from episode to episode. The pole length is
invisible to the agent and needs to be inferred from the response of the
system. In this task the sensor integration problem is thus paired with the need
for system identification. As can be seen in figure \ref{fig:pendulum}b, the RDPG
agent with an LSTM network reliably solves this task every time while a simple
feedforward agent (DDPG) fails entirely. RDPG with a simple RNN performs
considerably less well than the LSTM agent, presumably due to relatively long
episodes (T=350 steps) and the failure to backpropagate gradients effectively
through the plain RNN.  We found that a feedforward agent that does receive velocity
information can solve the variable-length swing-up task partly but does so less
reliably than the recurrent agent as it is unable to identify the relevant
system parameters (not shown).

\subsection{Memory tasks}

\begin{figure}
  \subfloat[][]{\includegraphics[width=4cm]{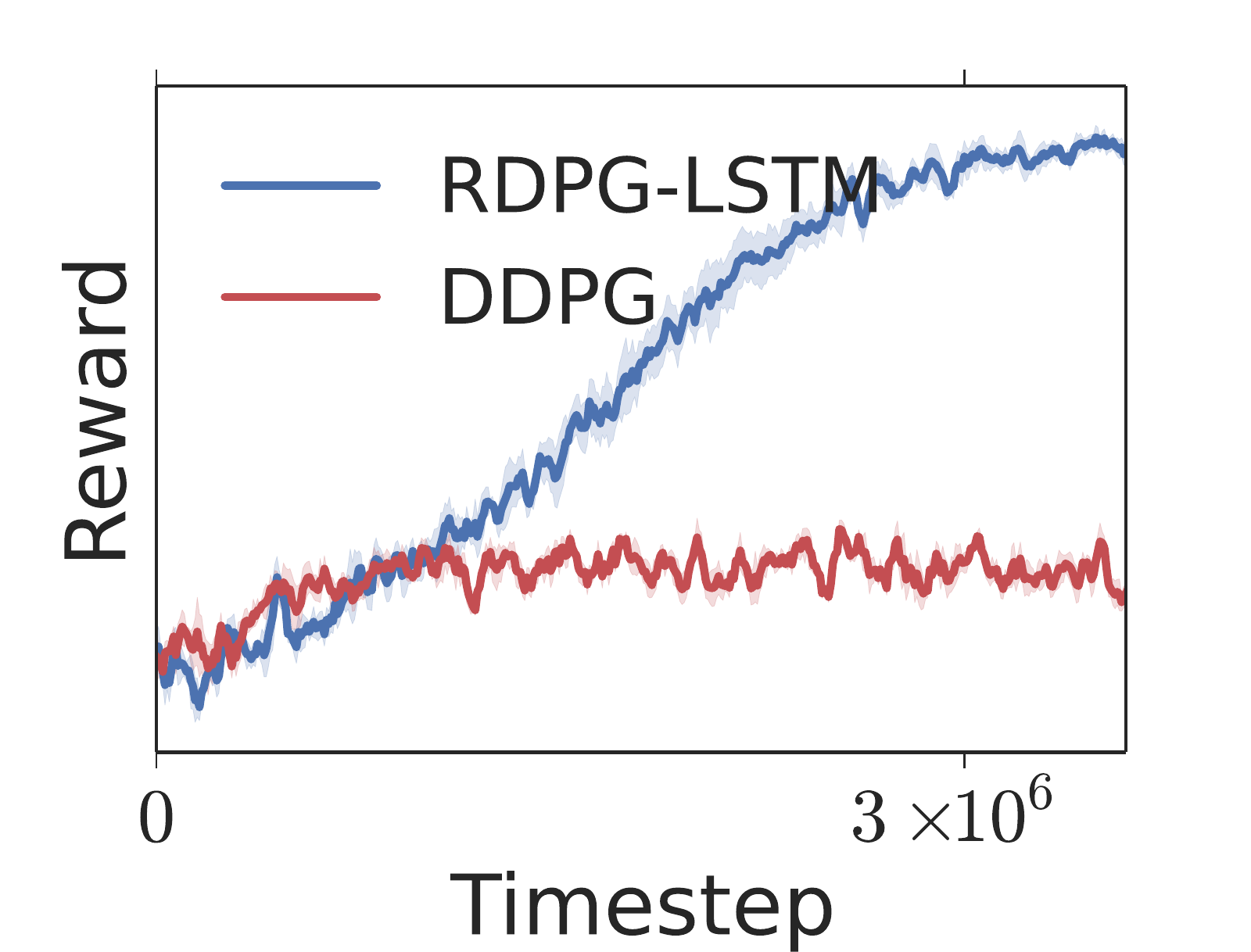} \label{sb:reacher}}
  \subfloat[][]{\includegraphics[width=4cm]{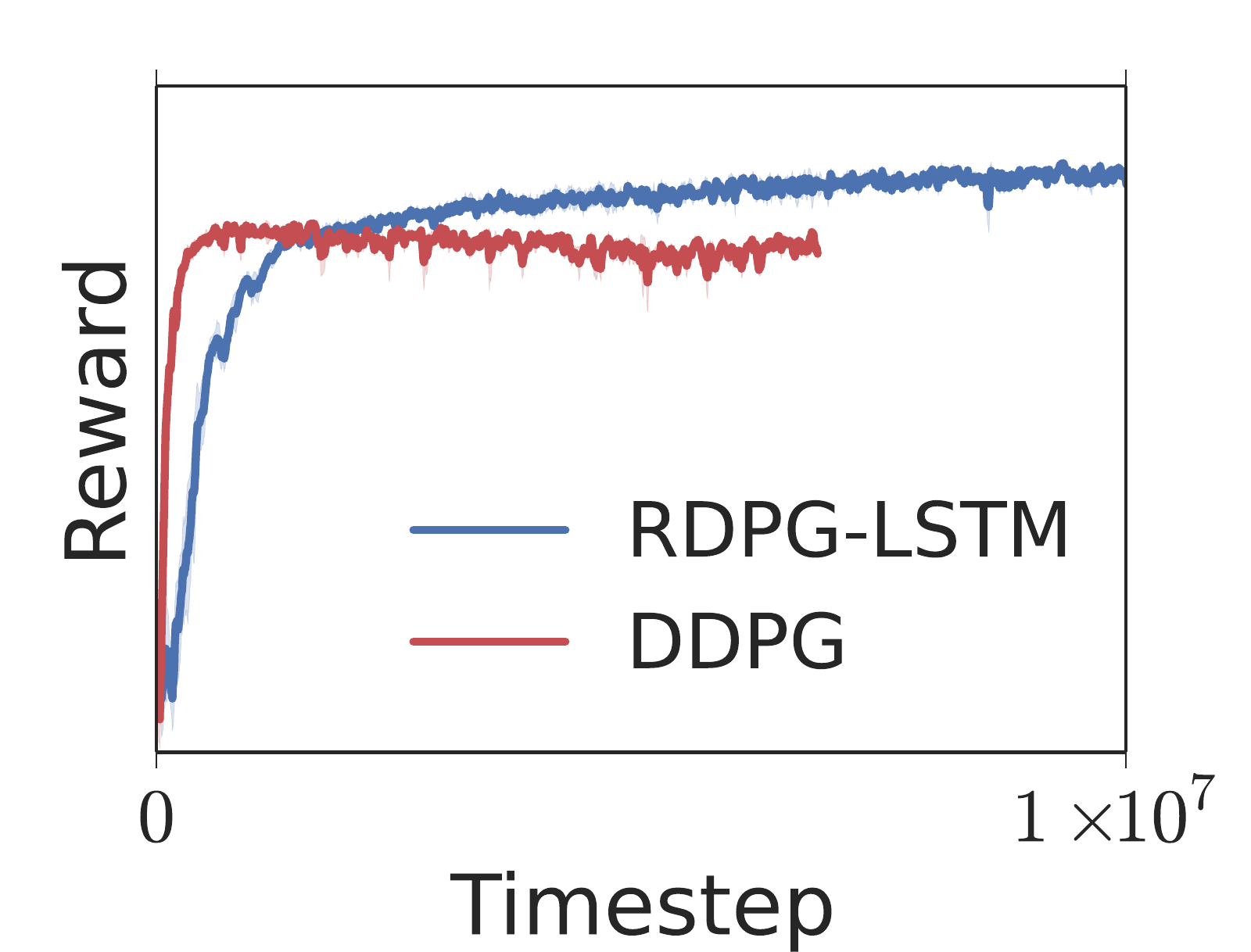} \label{sb:gripper}}
  \subfloat[][]{\includegraphics[width=3cm]{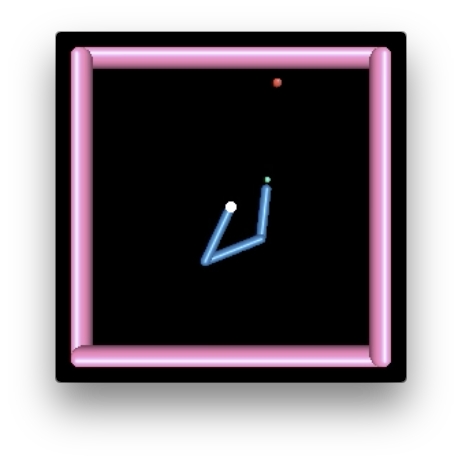} \label{sb:3ReacherExample}}
  \subfloat[][]{\includegraphics[width=3cm]{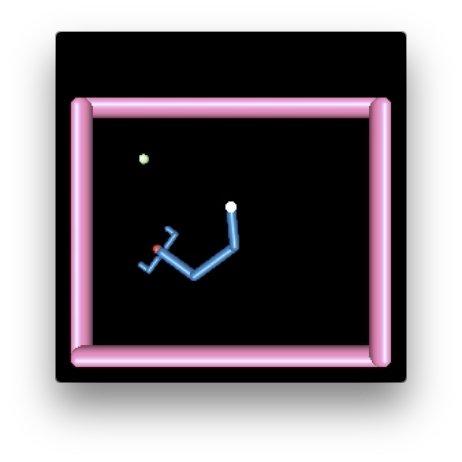} \label{sb:gripperExample}}
  \caption{ Reward curves for the \protect\subref{sb:reacher} hidden target reacher task, and
  \protect\subref{sb:gripper} return to start gripper task.
  In both cases the RDPG-agents with LSTMs are able to find good policies
  whereas the feedforward agents fail on the memory component. (In
  both cases the feedforward agents perform clearly better than random which is
  expected from the setup of the tasks: For instance, as can be seen in the
  video, the gripper without memory is still able to grab the payload and move
  it to a "default" position.)
  \\
  Example frames from the 3 joint reaching task \protect\subref{sb:3ReacherExample}
  and the gripper task \protect\subref{sb:gripperExample}.
  }
  \label{fig:memory}
\end{figure}

Another type of partially-observed task, which has been less studied in the
context of reinforcement learning, involves the need to remember explicit
information over a number of steps. We constructed two tasks like this. One was
a 3-joint reacher which must reach for a randomly positioned target, but the
position of the target is only provided to the agent in the initial
observation (the entire episode is 80 timesteps). As a harder variant of this task, we
constructed a 5 joint gripper which must reach for a (fully-observed) payload
from a randomized initial configuration and then return the payload to the
initial position of its "hand" (T=100). Note that this is a challenging control problem
even in the fully observed case. The results for both tasks are shown in figure
\ref{fig:memory}, RDPG agents with LSTM networks solve both tasks reliably
whereas purely feedforward agents fail on the memory components of the task as
can be seen in the supplemental video.

\subsection{Water maze}
\label{sec:Results:Watermaze}

The Morris water maze has been used extensively in rodents for the study of
memory \cite{dhooge2001applications}. We tested our algorithms on a
simplified version of the task. The agent moves in a 2-dimensional circular
space where a small region of the space is an invisible ``platform'' where the
agent receives a positive reward. At the beginning of the episode the agent and
platform are randomly positioned in the tank. The platform position is not
visible to the agent but it ``sees'' when it is on platform. The agent needs to
search for and stay on the platform to receive reward by controlling
its acceleration. After 5 steps on the
platform the agent is reset randomly to a new position in the tank but the
platform stays in place for the rest of the episode (T=200). The agent
needs to remember the position of the platform to return to it quickly.

\begin{figure}
  \subfloat[][]{\includegraphics[height=3.5cm]{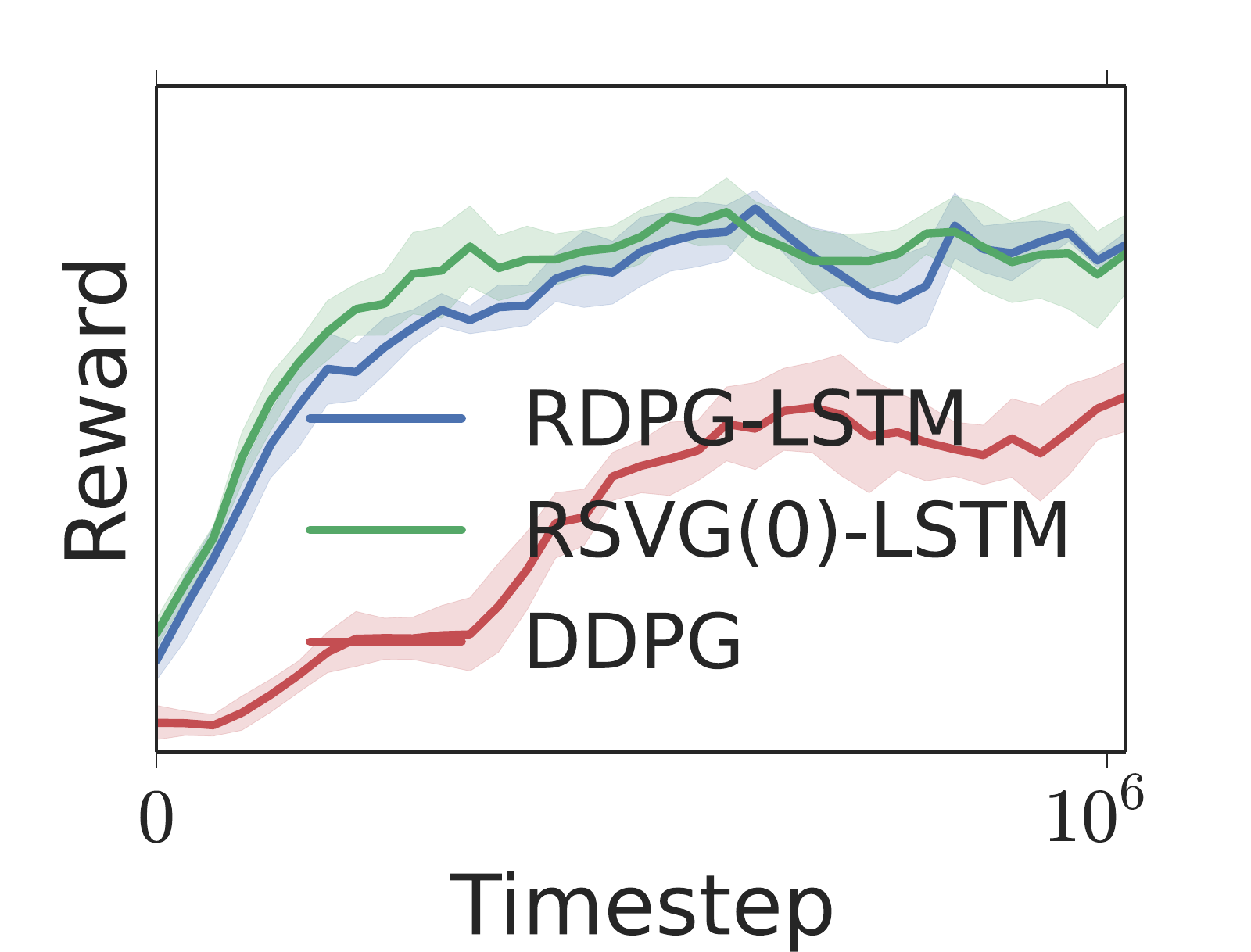} \label{sb:wmLearningCurve}}\hspace*{-1.5em}
  \subfloat[][]{\includegraphics[height=3.5cm]{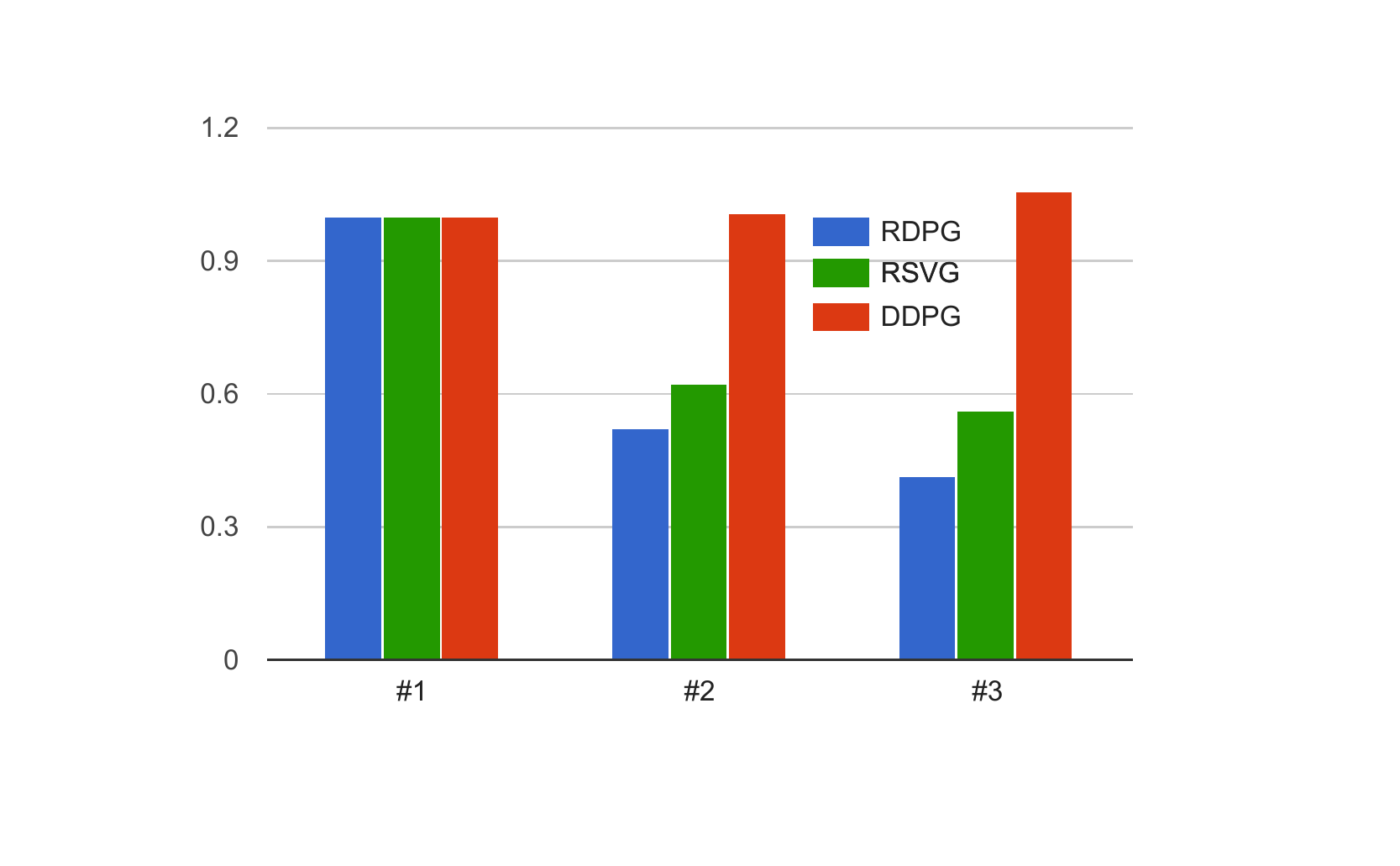} \label{sb:wmBar}}\hspace*{-3.5em}
  \subfloat[][]{\includegraphics[width=2cm]{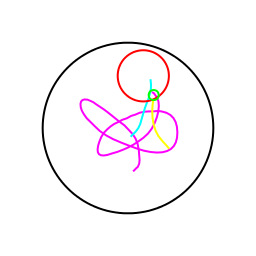} \label{sb:wmRDPG}}\hspace*{-1.2em}
  \subfloat[][]{\includegraphics[width=2cm]{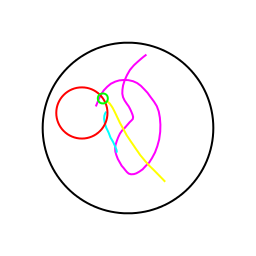} \label{sb:wmRSVG}}\hspace*{-1.2em}
  \subfloat[][]{\includegraphics[width=2cm]{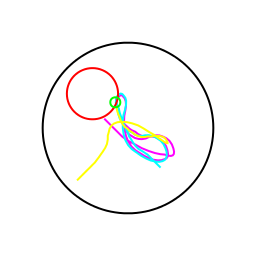} \label{sb:wmDDPG}}
  \caption{
  \protect\subref{sb:wmLearningCurve} shows the reward curve for different agents
  performing the water maze task. Both recurrent algorithms are capable of learning
  good solutions to the problem, while the non-recurrent agent (DDPG) is not.
  It is particularly notable that despite learning
  a deterministic policy, RDPG is able find search strategies that
  allow it to locate the platform.
  \protect\subref{sb:wmBar} This shows the number of steps the agents take
  to reach the platform after a reset, normalized by the number of steps taken for the
  first attempt. Note that on the 2nd and 3rd attempts the
  recurrent agents are able to reach the platform much more quickly, indicating they are
  learning to remember and recall the position of the platform.
  Example trajectories for the \protect\subref{sb:wmRDPG} RDPG, \protect\subref{sb:wmRSVG}
  RSVG(0) and \protect\subref{sb:wmDDPG} DDPG agents.
  Trajectory of the first attempt is purple,
  second is blue and third is yellow.
  }
  \label{fig:watermaze}
\end{figure}

It is sometimes presumed that a stochastic policy is required in order to solve
problems like this, which require learning a search strategy.
Although there is some variability in the results, we found that both RDPG
and RSVG(0) were able to find similarly
good solutions (figure \ref{sb:wmLearningCurve}), indicating RDPG is able to
learn reasonable, deterministic search strategies. Both solutions were able to
make use of memory to return to the platform more quickly after discovering it
during the initial search (figure \ref{sb:wmBar}). A non-recurrent agent (DDPG)
is able to learn a limited search strategy but fails to exploit memory to return the
platform after having been reset to a random position in the tank.

\subsection{High-dimensional observations}

We also tested our agents, with convolutional networks, on solving tasks
directly from high-dimensional pixel spaces. We tested on the pendulum
task (but now the agent is given only a static rendering of the pendulum at each timestep),
and a two-choice reaching task, where the target disappears after 5 frames
(and the agent is not allowed to move during the first 5 frames to prevent it from encoding the target position in its initial trajectory).

\begin{figure}
  \subfloat[][]{\includegraphics[width=4cm]{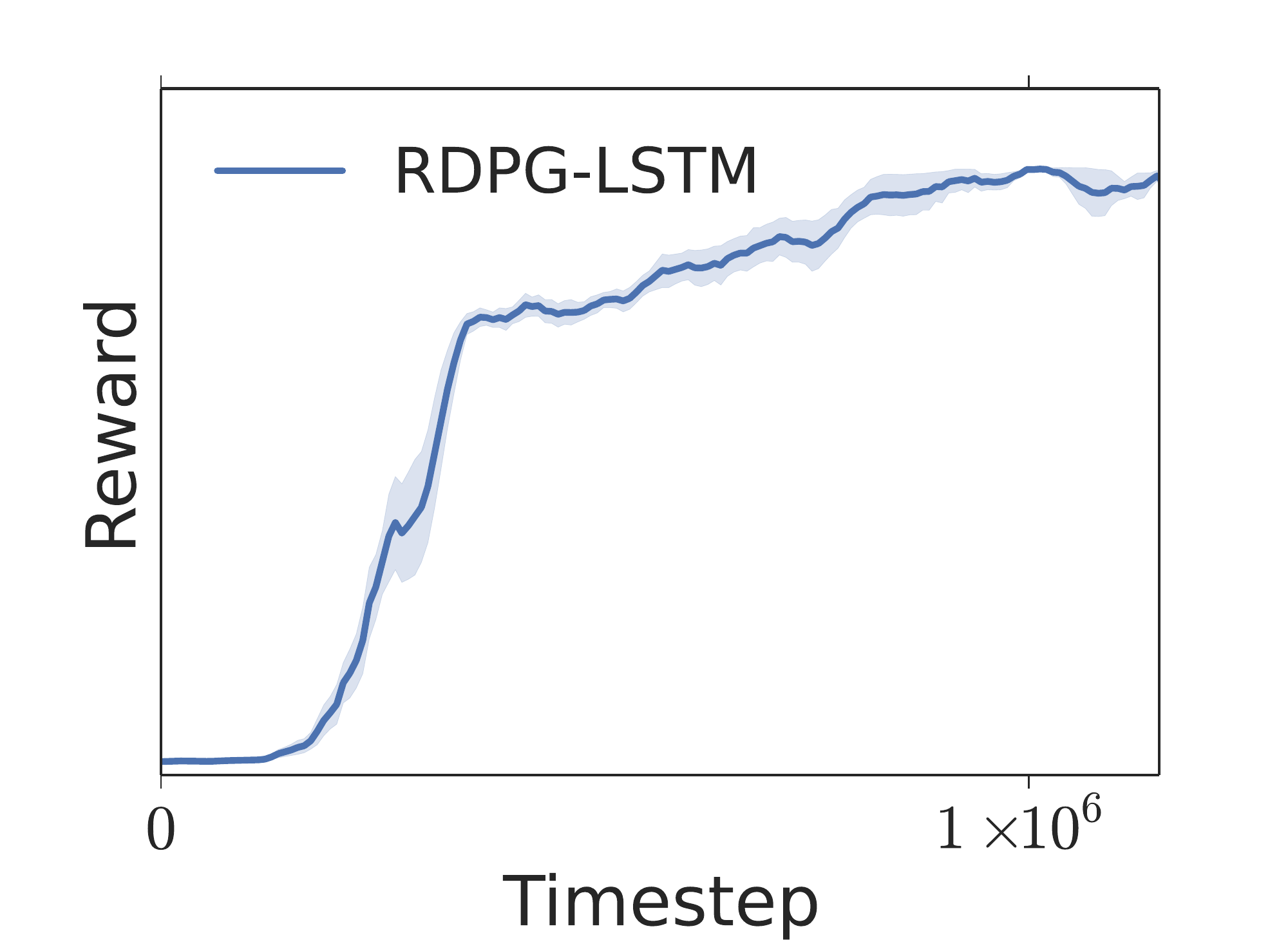} \label{sb:pixelsPendulum}}
  \subfloat[][]{\includegraphics[width=4cm]{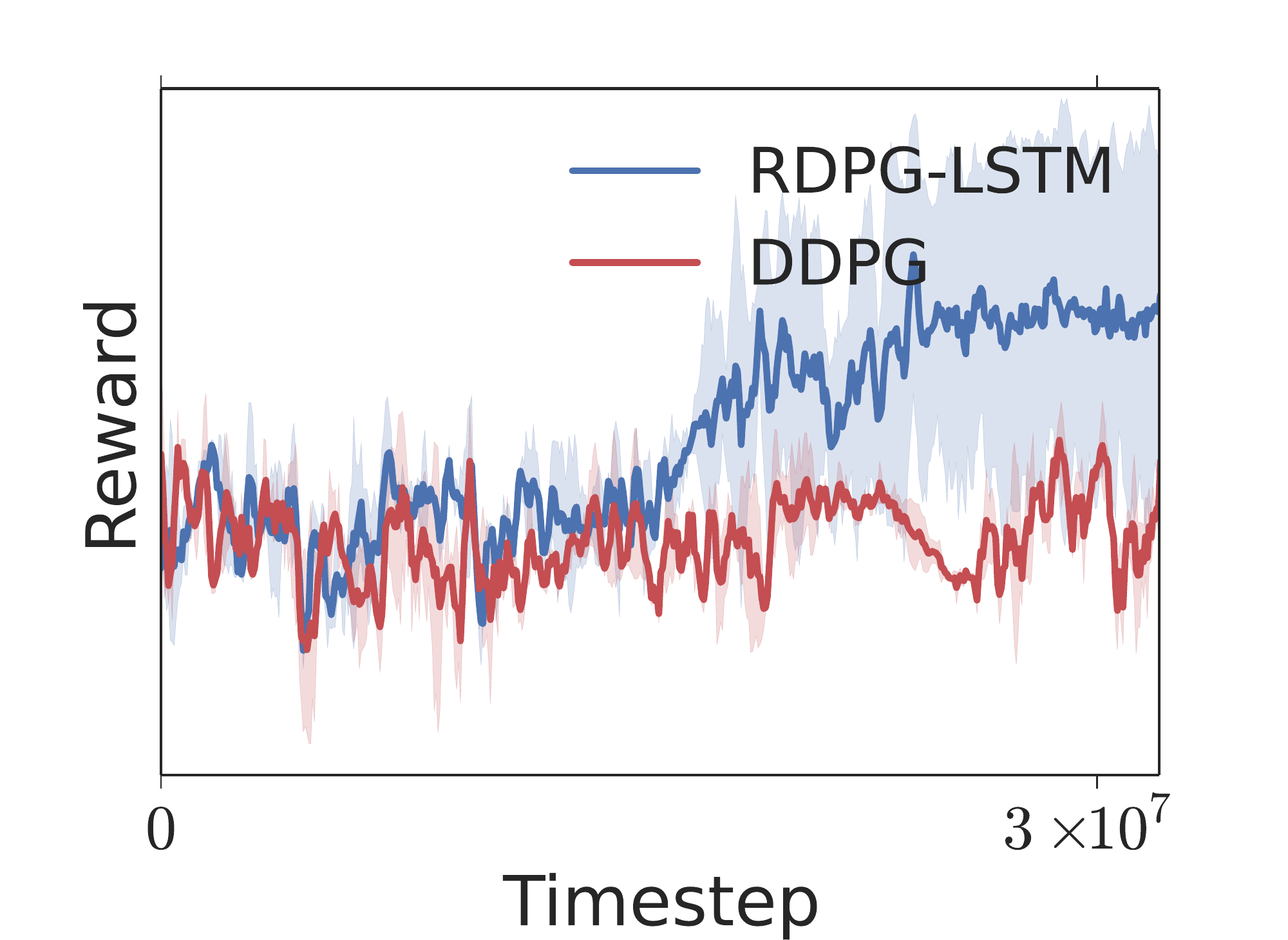} \label{sb:pixelsReacher}}
  \subfloat[][]{\includegraphics[width=3cm]{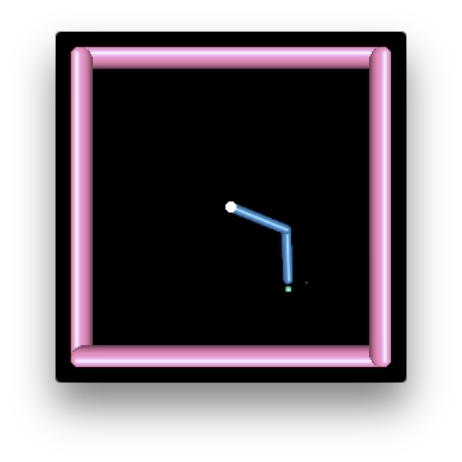} \label{sb:reacherPixelsExample}}
  \caption{
  RDPG was able to learn good policies directly from high-dimensional
  renderings for pendulum
  \protect\subref{sb:pixelsPendulum}, and a two choice reaching task with a disappearing target
  \protect\subref{sb:pixelsReacher}.
  \protect\subref{sb:reacherPixelsExample} Example frame from the reaching task.
  }
  \label{fig:pixels}
\end{figure}

We found that RDPG was able to learn effective policies from high-dimensional
observations which integrate information from multiple timesteps to estimate velocity and remember the
visually queued target
for the full length of the episode (in the reacher task). Figure \ref{fig:pixels}
shows the results.

\section{Discussion}

\subsection{Variants}

In the experiments presented here, the actor and critic networks are entirely
disjoint. However, particularly when learning deep, convolutional networks
the filters required in the early layers may be similar between the policy and
the actor. Sharing these early layers could improve computational efficiency
and learning speed.
Similar arguments apply to the recurrent part of the network,
which could be shared between the actor and the critic.
Such sharing, however, can also result in instabilities as updates to one network
may unknowingly damage or shift the other network. For this reason, we have not
used any sharing here, although it is a potential topic for further investigation.

\subsection{Related work}


There  is a large body of literature on solving partially observed control
problems. We focus on the most closely related work that aims to solve such problems with
learned memory.

Several groups \cite{lin1993hidden,bakker2002reinforcement,hausknecht2015deep}
 have studied the use of model-free algorithms with
 recurrent networks to solve POMDPs with discrete action spaces.
\cite{bakker2002reinforcement} focused on relatively long-horizon ("deep") memory
problems in small state-action spaces.
In contrast, \cite{hausknecht2015deep}
modified the Atari DQN architecture \cite{mnih2015human} (i.e.\ they perform
control from high-dimensional pixel inputs) and demonstrated that recurrent Q learning \cite{lin1993hidden}
 can perform the
required information integration to resolve short-term partial observability (e.g.\ to estimate velocities) that is
achieved via stacks of frames in the original DQN architecture.


Continuous action problems with relatively low-dimensional observation spaces
have been considered e.g.\ in
\cite{wierstra2007deep,wierstra2007critics,utsunomiya2009contextual,zhang2015memory}.
\cite{wierstra2007deep} trained LSTM-based stochastic policies using Reinforce;
\cite{wierstra2007critics,utsunomiya2009contextual,zhang2015memory} used
actor-critic architectures.
The algorithm of \cite{wierstra2007critics} can be seen as a special case of DPG
where the deterministic policy produces the
parameters of an action distribution from which the actions are then sampled.
This requires suitable exploration at the level of distribution parameters
(e.g.\ exploring in terms of means and variances of a Gaussian distribution); in
contrast, SVG(0) also learns stochastic policies but allows
exploration at the action level only.

All works mentioned above, except for \cite{zhang2015memory}, consider the memory
to be internal to the policy and learn the RNN parameters using BPTT,
back-propagating either TD errors or policy gradients. \cite{zhang2015memory}
instead take the view of \cite{peshkin1999external} and consider memory as extra
state dimensions that can can be read and set by the policy. They optimize the
policy using guided policy search \cite{levine2015end} which performs explicit
trajectory optimization along reference trajectories and, unlike our approach,
requires a well defined full latent state and access to this latent state during
training.

\section{Conclusion}

We have demonstrated that two related model-free approaches can be extended
to learn effectively with recurrent neural networks on a variety of partially-observed
problems, including directly from pixel observations. Since these algorithms
learn using standard backpropagation through time, we are able to benefit
from innovations in supervised recurrent neural networks, such as
long-short term memory networks \cite{hochreiter1997long}, to solve challenging
memory problems such as the Morris water maze.

\bibliographystyle{abbrv}

{
\bibliography{rdpg}}

\newpage

\section{Supplementary}

\begin{algorithm}[h]
  \caption{RSVG(0) algorithm \label{algo:rsvg0}}
  \begin{algorithmic}
    \STATE Initialize critic network $Q^\omega(a_t, h_t )$ and actor $\pi^\theta(h_t )$ with parameters $\omega$ and $\theta$.
    \STATE Initialize target networks $Q^{\omega'}$ and $\pi^{\theta'}$ with weights $\omega' \leftarrow \omega$, $\theta' \leftarrow \theta$.
    \STATE Initialize replay buffer $R$.
    \FOR{episodes = 1, M}
    	\STATE initialize empty history $h_0$
    	\FOR{t = 1, T}
		\STATE receive observation $o_t$
		\STATE $h_t \leftarrow h_{t-1}, a_{t-1}, o_t$ (append observation and previous action to history)
		\STATE select action $a_t = \pi^\theta(h_t, \nu) $ with $\nu \sim \beta$)
	\ENDFOR
      	\STATE Store the sequence $(o_1, a_1, r_1 ... o_T, a_T, r_T)$ in $R$
      	\STATE Sample a minibatch of $N$ episodes $(o_1^i, a_1^i, r_1^i, ... o_T^i, a_T^i, r_T^i)_{i=1, \dots ,N}$ from $R$
	\STATE Construct histories $h_t^i = (o_1^i, a_1^i, \dots a_{t-1}^i, o_t^i)$
        \STATE Compute target values for each sample episode
       	$(y_1^i, ... y_T^i)$ using the recurrent target networks
          \begin{equation*}
            y^i_t = r^i_t + \gamma Q^{\omega'}( h^i_{t+1},  \pi^{\theta'}(h^i_{t+1}, \nu) ) ~~~~\mathrm{with} ~~\nu \sim \beta
          \end{equation*}
      \STATE Compute critic update (using BPTT)
      \begin{equation*}
        \Delta \omega = \frac{1}{NT} \sum_i \sum_t (y^i_t - Q^\omega(h_t^i, a^i_t )\partialAt{Q^\omega(h_t^i, a^i_t )}{\omega}
      \end{equation*}

      \STATE Compute actor update (using BPTT)
      \begin{equation*}
        \Delta \theta = \frac{1}{NT} \sum_i \sum_t \partialAt{Q^\omega(h_t^i, \pi^\theta(h_t^i, \nu) )}{a}
        			\partialAt{\pi^\theta(h_t^i, \nu)}{\theta} ~~~~~~ \mathrm{with} ~~\nu \sim \beta
      \end{equation*}
      \STATE Update actor and critic using Adam \cite{kingma2014adam}
      \STATE Update the target networks
          \begin{align*}
            \omega' & \leftarrow \tau \omega + (1 - \tau) \omega' \\
            \theta'  & \leftarrow \tau \theta + (1 - \tau) \theta'
          \end{align*}
    \ENDFOR
  \end{algorithmic}
\end{algorithm}


\end{document}